%% file: main.tex
\def\year{2022}\relax
\documentclass[letterpaper]{article} 
\usepackage{aaai22}  
\usepackage{times}  
\usepackage{helvet}  
\usepackage{courier}  
\usepackage[hyphens]{url}  
\usepackage{graphicx} 
\usepackage{natbib}  
\usepackage{caption} 
\DeclareCaptionStyle{ruled}{labelfont=normalfont,labelsep=colon,strut=off} 
\usepackage[table,dvipsnames]{xcolor}
\usepackage{pgfplots}
\usepackage{latexsym}
\usepackage{dsfont}
\usepackage{color}
\usepackage{booktabs} 
\usepackage{subfigure}
\usepackage{CJKutf8}
\usepackage{graphicx}
\usepackage{amsmath}
\usepackage{multirow, makecell, caption}
\usepackage{floatrow}
\newfloatcommand{capbtabbox}{table}[][\FBwidth]

\usepackage[normalem]{ulem}
\usepackage{amssymb}
\usepackage{amsfonts}
\usepackage{multicol}
\usepackage{tipa}
\usepackage{bm}
\usepackage[ruled,linesnumbered]{algorithm2e}

\usepackage{paralist}
\usepackage{IEEEtrantools}
\usepackage[switch]{lineno}

\newtheorem{definition}{Definition}

\newcommand{\method}{\textsc{LoReN}\xspace}

\title{\method: Logic-Regularized Reasoning for Interpretable Fact Verification}

\author {
    Jiangjie Chen\textsuperscript{\rm $\spadesuit\clubsuit$}\thanks{Work is done during internship at ByteDance AI Lab.},
    Qiaoben Bao\textsuperscript{\rm $\spadesuit$}, 
    Changzhi Sun\textsuperscript{\rm $\clubsuit$},
    Xinbo Zhang\textsuperscript{\rm $\clubsuit$},\\
    Jiaze Chen\textsuperscript{\rm $\clubsuit$},
    Hao Zhou\textsuperscript{\rm $\clubsuit$},
    Yanghua Xiao\textsuperscript{\rm $\spadesuit\diamondsuit$}\thanks{Corresponding authors.},
    Lei Li\textsuperscript{\rm $\heartsuit$}\footnotemark[2]\thanks{Work is done while at ByteDance AI Lab.}
}
\affiliations {
    \textsuperscript{\rm $\spadesuit$}Shanghai Key Laboratory of Data Science, School of Computer Science, Fudan University\\
    \textsuperscript{\rm $\clubsuit$}ByteDance AI Lab
    \textsuperscript{\rm $\heartsuit$}University of California, Santa Barbara\\
    \textsuperscript{\rm $\diamondsuit$}Fudan-Aishu Cognitive Intelligence Joint Research Center\\
    \texttt{\{jjchen19, qbbao19, shawyh\}@fudan.edu.cn}, \ \ 
    \texttt{lilei@cs.ucsb.edu}, \\
    \texttt{\{sunchangzhi, zhangxinbo.freya, chenjiaze, zhouhao.nlp\}@bytedance.com}\\
}

\begin{document}
\maketitle

\begin{abstract}
\input{000abstract}
\end{abstract}

\section{Introduction}
\label{sec:intro}

\input{010intro}

\section{Related Work}
\label{sec:related}
\input{015related}

\section{Proposed Approach}
\label{sec:method}
\input{020method}

\section{Experiments}
\label{sec:experiment}
\input{030experiment}

\section{Results and Discussion}
\label{sec:results}
\input{040results}

\section{Conclusion and Future Work}
\label{sec:conclusion}
\input{060conclusion}

\section*{Acknowledgements}
We thank Rong Ye, Jingjing Xu and other colleagues at ByteDance AI Lab as well as the anonymous reviewers for the discussions and suggestions for the manuscript.
This work was supported by National Key Research and Development Project (No. 2020AAA0109302), Shanghai Science and Technology Innovation Action Plan (No.19511120400) and Shanghai Municipal Science and Technology Major Project (No.2021SHZDZX0103).

\bibliography{aaai22}

\clearpage
\appendix
\label{sec:supplementary}

\input{070supplementary}

\end{document}

%% file: 000abstract.tex
Given a natural language statement, how to verify its veracity against a large-scale textual knowledge source like Wikipedia?
Most existing neural models make predictions without giving clues about which part of a false claim goes wrong.
In this paper, we propose \method, an approach for interpretable fact verification.
We decompose the verification of the whole claim at phrase-level, where the veracity of the phrases serves as explanations and can be aggregated into the final verdict according to logical rules.
The key insight of \method is to represent claim phrase veracity as three-valued latent variables, which are regularized by aggregation logical rules.
The final claim verification is based on all latent variables.
Thus, \method enjoys the additional benefit of interpretability --- it is easy to explain how it reaches certain results with claim phrase veracity. 
Experiments on a public fact verification benchmark show that \method is competitive against previous approaches while enjoying the merit of faithful and accurate interpretability.
The resources of \method are available at: \url{https://github.com/jiangjiechen/LOREN}.

%% file: 010intro.tex
\begin{figure}[t]
    \centering
	\includegraphics[width=\linewidth]{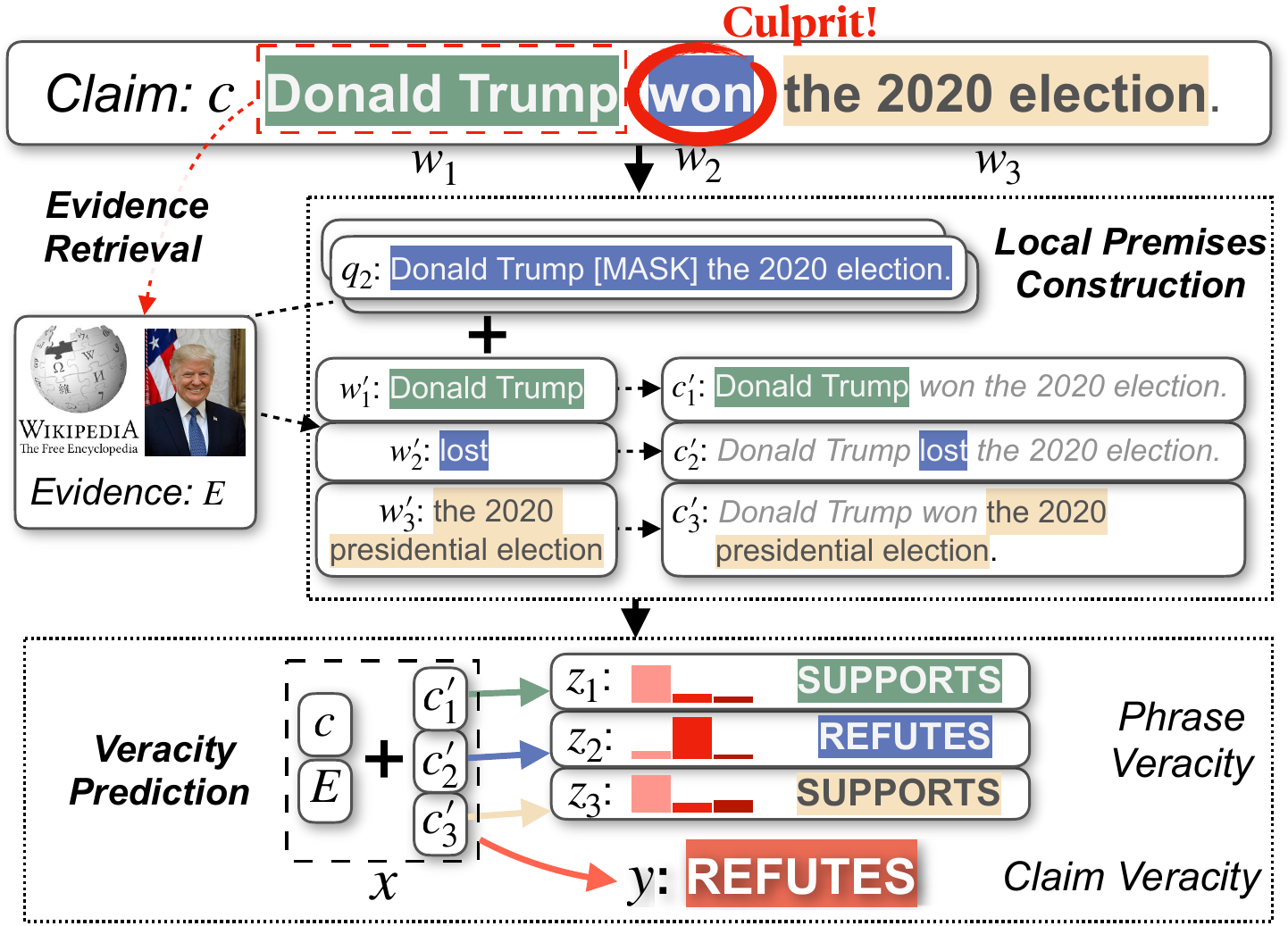}
    \caption{An example of how our proposed fact verification framework \method works. Texts highlighted denote the data flow for three phrases extracted from the claim. 
    \method not only makes the final verification but also finds the culprit phrase ($w_2$) that causes the claim's falsity. 
    }
    \label{fig:front}
\end{figure}

The rapid growth of mobile platforms has facilitated creating and spreading of information.
However, there are many dubious statements appearing on social media platforms.
For example, during the 2020 U.S. presidential election, there are many false claims about Donald Trump winning the election, as shown in Figure \ref{fig:front}.
Verifying these statements is in critical need.
How to verify the validity of a textual statement?
We attempt to predict whether a statement is \textit{supported}, \textit{refuted} or \textit{unverifiable} with an additional large textual knowledge source such as Wikipedia.
Notice that it is computationally expensive to compute the input statement with every sentence in Wikipedia.

This work focuses on interpretable fact verification --- it aims to provide decomposable justifications in addition to an overall veracity prediction.
We are motivated by a simple intuition: the veracity of a claim depends on the truthfulness of its composing phrases, e.g., subject, verb, object phrases.
A false claim can be attributed to one or more unsupported phrases, which we refer to as the \emph{culprit}.
The claim is valid if all phrases are supported by certain evidence sentences in Wikipedia.
For example, one culprit in Figure \ref{fig:front} would be the phrase ``won''.
Therefore, faithful predictions of phrase veracity would explain why a verification model draws such a verdict.
In addition, through phrasal veracity prediction, identifying the culprit also alleviates the burden of correcting an untrustworthy claim, as we can easily alter ``won'' to ``lost'' to make it right.

Most current studies focus on designing specialized neural network architectures, with the hope of exploiting the semantics from sentences \cite{Nie2019CombiningFE,zhou-etal-2019-gear,liu2020fine,zhong2020reasoning,si-etal-2021-topic,jiang-etal-2021-exploring-listwise}.
However, these methods are limited in interpretability, as they usually only give an overall verdict.
This puts forth trust issues for humans, as a decision is usually made in a black-box fashion.
Recent studies on explainable fact verification \cite{stammbach2020fever,samarinas-etal-2021-improving,wu-etal-2021-unified} mostly focus on giving intuitive display of the key basis for the model results, instead of building an interpretable models that output the reasons for obtaining the results while giving the results.

It is challenging to learn interpretable models that not only predict but also explain its rationale, since there is a lack of truthfulness labels on the phrase level.
There are public datasets with veracity label for the overall claim, but it is unknown which part of the claim makes it untrustworthy.
Manually annotating such fine-grained data is unrealistic and requires tremendous human labor.
How to supervise a model to reach meaningful phrasal veracity predictions?
Our insight comes from the intuition that simple \textit{symbolic logical rules} can be utilized to create weak supervisions for intermediate phrasal predictions.
Empirically, all phrases should be supported if a claim is true, and a claim is refuted if there exists at least one false phrase.
If the outcome of a claim is \textit{unverifiable}, then there must be no \textit{refuted} phrase and at least one phrase that should be verified as \textit{unverifiable}.
With the logical rules in mind, we only have to identify the patterns these suspicious phrases give during training.

For this purpose, we propose \method, a \textsc{Lo}gic-\textsc{Re}gularized \textsc{N}eural latent model, to predict the veracity of a claim, as well as to give explanation.
The overall idea of \method is to decompose the verification into a series of hypothesis verification at the phrase level, which are constrained by the introduced logical aggregation rules.
Each rule concerns the compositional logic that describes how phrasal veracity predictions are logically aggregated into claim veracity.
Together, the veracity prediction of every claim phrase serves as the atomic propositions of the compositional logic.
Thus, a key perspective of \method is to represent these phrase veracity as latent variables regularized by the softened aggregation logic for meaningful predictions for claim phrases.
To solve this latent model, \method adopts a modern approach using amortized variational inference (variational auto-encoding) \cite{DBLP:journals/corr/KingmaW13}.
Furthermore, \method constructs the teacher model by aggregating logic over all latent variables, distilling logical knowledge to the student (claim verification) model.
To arrive at these propositions, as another key perspective, we convert the problem of finding relevant phrases in evidence into a machine reading comprehension (MRC) task, where we generate probing questions for evidence to answer.
To summarize, the contributions of this work include:
\begin{itemize}
    \item We propose an interpretable method \method to predict the veracity of both a claim sentence and its phrases.
    \item We present a technique to weakly supervise phrasal veracity learning with a MRC module and latent variable modeling regularized by logical rules.
    \item We experiment \method on FEVER \cite{thorne-etal-2018-fever}, a large fact verification benchmark. Besides competitive verification results, \method also provides \textit{faithful} (over 96\% agreement) and \textit{accurate} phrasal veracity predictions as explanations.
\end{itemize}

%% file: 015related.tex
There are several related problems about verifying the truthfulness of one or multiple sentences, including natural language inference (NLI) \cite{kang18acl}, claim verification \cite{thorne-etal-2018-fever}, misinformation detection \cite{zellers2019defending}, etc.
In this paper, we study the claim verification task \cite{thorne-etal-2018-fever}, which focuses on verifying claims against trustworthy knowledge sources.
The majority of existing studies adopt a two-step pipeline to verify a textual claim, i.e., evidence retrieval and claim verification.
Current verification systems can be categorized by the granularity of the interaction between claim and evidence, including those of sentence-level \cite{Nie2019CombiningFE,zhou-etal-2019-gear}, semantic role-level \cite{zhong2020reasoning} and word-level \cite{liu2020fine}.
They learn the representations of claim and evidence sentences of different granularity based on neural networks and gives a final verdict in an end-to-end fashion.
In contrast, we conduct \textit{phrase-level} verification and take a further step forward to more interpretable reasoning and verification.

There are some recent studies on interpretable fact verification, such as using GPT-3 \cite{NEURIPS2020_1457c0d6} to summarize evidence and generate explanations \cite{stammbach2020fever}, pointing out salient pieces in evidence with attention weights \cite{samarinas-etal-2021-improving}, and picking relevant sentences in retrieved evidence \cite{wu-etal-2021-unified}.
Instead, we take a different route towards interpretable fact verification by producing where and how a claim is falsified.
The final verdict is drawn based on explanations, making a step forward to being right for the right reasons.

Previous efforts towards unifying symbolic logic and neural networks include those of \citet{sourek2015lifted,manhaeve2018deepproblog,lamb2020graph}.
A class of integrated symbolic logic and neural network methods is based on the variational EM framework \cite{qu2019probabilistic,DBLP:conf/nips/ZhouHZLSXT20}.
Another standard method is to soften logic with neural network components \cite{hu-etal-2016-harnessing,li-etal-2019-logic,wang2020integrating}, which can be trained in an end-to-end manner.
Our method draws inspiration from both lines of work.
We represent the intermediary veracity predictions as latent variables in latent space, which are regularized with softened logic.



%% file: 020method.tex
In this section, we present the proposed method \method for verifying a textual claim against a trustworthy knowledge source (e.g., Wikipedia), which consists of two sub-tasks: \textit{1)} evidence retrieval and \textit{2)} fact verification.
In this paper, we primarily focus on fact verification and assume evidence text (e.g., several related sentences) is retrieved by a separate method.
A possible verification result can be \textit{supported} (\texttt{SUP}), \textit{refuted} (\texttt{REF}) or \textit{not-enough-information} (\texttt{NEI}).

Different from most previous methods that give an overall prediction, our goal is to predict the final \textit{claim veracity} and faithful \textit{phrase veracity} as explanations.
First, we define the task of \textit{claim verification} and \textit{phrase verification}.

\paragraph{Claim Verification}

Given a claim sentence $c$ and retrieved evidence text $E$, our goal is to model the probability distribution  $p(\bm{y}| c, E)$, where $\bm{y} \in \{\texttt{SUP}, \texttt{REF},\texttt{NEI} \}$ is a three-valued variable indicating the veracity of the claim given evidence.
In this paper, $\bm{bold}$ letters indicate variables.

\paragraph{Phrase Verification}
We decompose the verification of a claim at phrase-level, and predict the veracity $\bm{z}_i$ of a claim phrase $w_i\in \mathcal{W}_c$ by $p(\bm{z}_i|c, w_i, E)$, where $\bm{z}_i \in \{\texttt{SUP}, \texttt{REF}, \texttt{NEI}\}$. 
We extract the claim phrases $\mathcal{W}_c$ with a set of heuristic rules using a series of off-the-shelf tools provided by AllenNLP \cite{Gardner2017AllenNLP}.
Claim phrases include named entities (NEs), verbs, adjective and noun phrases (APs and NPs).

Specifically, we leverage a part-of-speech (POS) tagger for identifying verbs and a constituency parser to identify noun phrases (NPs).
For the fine-grained extraction of NPs, we further decompose them into more fine-grained ones using POS tagger and named entity recognizer (NER). 
We use several simple \textit{heuristic} rules, including: 
1) we parse all leaf NPs, and keep all verbs with a POS tagger;
2) we break down the NPs with an NER and isolate the adjectives from NPs for finer-grained phrases.
For example, we have ``\textit{Donald Trump}'' (NE), ``\textit{won}'' (verb) and ``\textit{the 2020 election}'' (NP) as claim phrases in Figure \ref{fig:front}.

\subsection{Logical Constraints}
\label{sec:logic}
After introducing the phrase verification, we observe that some natural, logical consistencies between phrase verification and claim verification should be satisfied. Specifically, a claim is found 
\begin{inparaenum}[\it 1)]
    \item \texttt{REF} if at least one claim phrase is refuted by evidence;
    \item \texttt{SUP} if all claim phrases are supported; 
    \item \texttt{NEI} if neither of the above, that is, there is no contradictory but at least one phrase gets unknown outcome.
\end{inparaenum}
Notice that the checking rule for the \texttt{REF} judgment has priority over \texttt{NEI}, because it is also possible for a phrase to be \texttt{NEI} in a \textit{refuted} claim, but not vice versa.
Formally, we give the following definition of the aggregation logic.
\begin{definition}
\label{def:logic}
Given a statement $c$, a set of claim phrases $\mathcal{W}_c$, and a set of evidence $E$, with $\top(c)$, $\bot(c)$ and $\circleddash(c)$ denoted as true, false and unknown respectively.
$V(c, \mathcal{W}_c, E)$ is defined as the value of $c$ taking one of the three, i.e. $\{\top, \bot, \circleddash\}$ w.r.t. $\mathcal{W}_c$ given evidence $E$, which corresponds to the predicted label $y \in \{\texttt{SUP}, \texttt{REF}, \texttt{NEI}\}$.
Then we have:
\begin{equation*}
\begin{split}
    V(c, \mathcal{W}_c, E) \models \top, \ \mathrm{iff} \ \ &\forall w \in \mathcal{W}_c, V(c, w, E) \models \top \\
    V(c, \mathcal{W}_c, E) \models \bot, \  \mathrm{iff} \ \  &\exists w \in \mathcal{W}_c, V(c, w, E)\models \bot \\
    V(c, \mathcal{W}_c, E) \models \circleddash,\  \mathrm{iff} \ \  &\neg (V(c, \mathcal{W}_c, E) \models \top) \wedge \\
    &\forall w \in \mathcal{W}_c, V(c,w,E)\models \{\top, \circleddash\}
    \label{eq:logic}
\end{split}
\end{equation*}
where $V(c, w, E)$ is defined as the value of $c$ w.r.t. a single claim phrase $w$ and the given evidence $E$.
\end{definition}

With the logic in mind, we then introduce how \method learns to predict the veracity of both a claim and its phrases without direct supervision for the latter.

\subsection{Overview of \method}

The basic idea of \method is to decompose the verification of a claim at phrase-level, and treats the veracity of each phrase $w_i \in \mathcal{W}_c$ as a three-valued latent variable $\bm{z}_i$. 
We define $\bm{z} = (\bm{z}_1, \bm{z}_2, \dots, \bm{z}_{|\mathcal{W}_c|})$.
The veracity of a claim $\bm{y}$ depends on the latent variables $\bm{z}$.
Inspired by \citet{hu-etal-2016-harnessing}, to impose the logical constraints mentioned above,
we propose a distillation method that transfers the logical knowledge into the latent model.
Next, we will detail the latent model and the logical knowledge distillation.

\subsubsection{Latent Model}
We formulate the fact verification task in a probabilistic way.
Given an input $x = (c, E)$ consisting of textual claim $c$ and retrieved evidence text $E$, we define target distribution $p_\theta(\bm{y}| x)$ as below:
\begin{equation}
    p_{\theta}(\bm{y} | x) = \sum_{\bm{z}} p_\theta(\bm{y} | \bm{z}, x) p(\bm{z} | x)
\end{equation}
where $p(\bm{z}|x)$ is the prior distribution over latent variable $\bm{z}$ conditioned on the input $x$, and $p_\theta$ gives the probability of $\bm{y}$ conditioned on $x$ and latent $\bm{z}$.
Note that we assume that $\bm{z}_i$ is independent of each other, namely, $p(\bm{z}|x) = \prod_i p(\bm{z}_i | x, w_i)$.
Given the gold label $y^\star$, the objective function is to minimize the negative likelihood as follow:
\begin{equation}
    \mathcal{L}(\theta) = -\log p_\theta(y^\star | x).
\end{equation}
Theoretically, we can adopt the EM algorithm for optimization.
However, in our setting, it is difficult to compute the exact posterior $p_\theta(\bm{z} | \bm{y}, x)$ due to the large space of $\bm{z}$.
With recent advances in the variational inference \cite{DBLP:journals/corr/KingmaW13}, we could amortize the variational posterior distribution with neural networks.
It results in the well-known variational bound (negative Evidence Lower BOund, ELBO) to be minimized:
\begin{equation}
     \overbrace{\mathop{-\mathbb{E}}_{q_\phi(\bm{z} | \bm{y}, x)} [\log  p_\theta(y^*|\bm{z}, x))] + D_\mathrm{KL}(q_\phi(\bm{z}|\bm{y}, x) \parallel p(\bm{z} | x))}^{\text{negative ELBO} : \mathcal{L}_\mathrm{var}(\theta, \phi)}
\end{equation}
where $q_\phi(\cdot)$ is the variational posterior distribution conditioned on $\bm{y}, x$, and $D_\mathrm{KL}$ is Kullback–Leibler divergence.
In experiments, we use an off-the-shelf and pre-trained NLI model as prior distribution $p(\bm{z}|x)$, whose parameters are fixed.\footnote{We use a DeBERTa \cite{he2021deberta} fine-tuned on MNLI dataset \cite{bowman-etal-2015-large} as the NLI model.}
The NLI model yields the distribution of \texttt{contradicted}, \texttt{neutral} and \texttt{entailment}, which we take correspond to \texttt{REF}, \texttt{NEI} and \texttt{SUP} to some extent.

\subsubsection{Logical Knowledge Distillation}

To integrate the information encoded in the logical rules into latent variables,
we propose a distillation method, 
which consists of a \textit{teacher model} and a \textit{student model}.
The student model is the $p_\theta(\bm{y}| \bm{z}, x)$ we intend to optimize.
The teacher model is constructed by projecting variational distribution $q_\phi(\bm{z}|\bm{y}, x)$ into a subspace, denoted as $q_\phi^\mathrm{T}(\bm{y}_z|\bm{y}, x)$.
The subspace is constrained by the logical rules, since $\bm{y}_z$ is the logical aggregation of $\bm{z}$.
Thus, simulating the outputs of $q_\phi^\mathrm{T}$ serves to transfer logical knowledge into $p_\theta$.
Formally, the distillation loss is formulated as:
\begin{equation}
\label{eq:logicloss}
    \mathcal{L}_\mathrm{logic}(\theta, \phi) = D_\mathrm{KL}\left(  p_\theta(\bm{y} | \bm{z}, x) \parallel q_\phi^\mathrm{T}(\bm{y}_z | \bm{y}, x)\right).
\end{equation}

Overall, the final loss function is defined as the weighted sum of two objectives:
\begin{equation}
\label{eq:loss}
    \mathcal{L}_\mathrm{final}(\theta, \phi) = (1 - \lambda )\mathcal{L}_\mathrm{var}(\theta, \phi) + \lambda \mathcal{L}_\mathrm{logic}(\theta, \phi)
\end{equation}
where $\lambda$ is a hyper-parameter calibrating the relative importance of the two objectives.

\subsection{Teacher Model Construction}
\label{sec:prior}
ELBO cannot guarantee latent variables to be the veracity of corresponding claim phrases without any direct intermediate supervisions.
As a key perspective, they are aggregated following previously described logical rules, making them weak supervisions for phrase veracity.

To this end, we relax the logic with soft logic \cite{li-etal-2019-logic} by product t-norms for differentiability in training and regularization of latent variables.
According to $\mathsection$\ref{sec:logic}, given probability of the claim phrase veracity $\bm{z}$, we logically aggregate them into $\bm{y}_z$ as follows (for simplicity, we drop the input $x$):
\begin{equation}
\begin{split}
    q_\phi^\mathrm{T}(\bm{y}_z = \texttt{SUP}) &= \mathop{\Pi}_{i=1}^{|\bm{z}|} q_\phi(\bm{z}_i = \texttt{SUP}) \\
    q_\phi^\mathrm{T}(\bm{y}_z = \texttt{REF}) &= 1 - \mathop{\Pi}_{i=1}^{|\bm{z}|} \left(1 - q_\phi(\bm{z}_i = \texttt{REF})\right) \\
    q_\phi^\mathrm{T}(\bm{y}_z = \texttt{NEI}) &= 1 - q_\phi^\mathrm{T}(\bm{y}_z = \texttt{SUP}) - q_\phi^\mathrm{T}(\bm{y}_z = \texttt{REF}) 
\label{eq:softlogic}
\end{split}
\end{equation}
where $\sum_{\bm{y}_z} q_\phi^\mathrm{T}(\bm{y}_z) = 1$ and $\sum_{\bm{z}_i} q_\phi (\bm{z}_i) = 1$.

The prediction behavior of $q_\phi^\mathrm{T}$ reveals the information of the rule-regularized subspace, indicating the uncertain and probabilistic nature of the prediction \cite{chen-etal-2020-uncertain}.
By minimizing the distillation loss $\mathcal{L}_\mathrm{logic}$ in Eq. \ref{eq:logicloss}, the phrasal veracity predictions are regularized by the aggregation logic even if we do not have specific supervisions for claim phrases.

\subsection{Building Local Premises}

Before parameterizing $p_\theta(\cdot)$ and $q_\phi(\cdot)$ in the latent model, we find the information required for verifying each claim phrase from evidence in an MRC style.
We collect them into a set of \textit{local premises} corresponding to each claim phrase, which is important for \method's interpretability w.r.t. phrasal veracity.
One of the key perspective is to convert the finding of such information into a generative machine reading comprehension (MRC) task, which requires a question generation and answering pipeline.

\subsubsection{Probing Question Generation}

Before MRC, we first build probing questions $\mathcal{Q}$ for every claim phrase respectively.
Each question consists of two sub-questions: one \textit{cloze} questions \cite{devlin-etal-2019-bert} (e.g., ``\textit{\texttt{[MASK]} won the 2020 election.}'') and \textit{interrogative} questions \cite{wang2020neural} (e.g., ``\textit{Who won the 2020 election?}'').
Both types of questions are complementary to each other.
The cloze questions lose the semantic information during the removal of masked phrases (e.g., ``\textit{he was born in \texttt{[MASK]}}'', where \textit{\texttt{[MASK]}} can either be a place or a year.).
And the generated interrogative ones suffer from the incapability of a text generator.
In experiments, we use an off-the-shelf question generation model based on $\text{T5}_{base}$~\cite{raffel2020exploring} to generate interrogative questions.

\subsubsection{Local Premise Construction}

For every claim phrase $w_i \in \mathcal{W}_c$, we first generate probing question $q_i \in \mathcal{Q}$ with off-the-shelf question generators.
The MRC model takes as an input $\mathcal{Q}$ and $E$ and answers $\mathcal{W}_E$.
Then, we replace $w_i \in \mathcal{W}_c$ with answers $w'_i \in \mathcal{W}_E$, yielding replaced claims $c'_i$ such as ``\textit{Donald Trump \uline{\textit{lost}} the 2020 election}'', where $w'_i=$``\textit{lost}'' and $w_i=$``\textit{won}''.
Such replaced claims are denoted as local premises $\{c'_i\}_{i=1}^{|\mathcal{W}_c|}$ to reason about the veracity of every claim phrase.

\subsubsection{Self-supervised Training of MRC}

The MRC model is fine-tuned in a self-supervised way to adapt to this task at hand.
The MRC model takes as input a probing question and evidence sentences and outputs answer(s) for the question.
During training, claim phrases $\mathcal{W}_c$ in a claim are used as ground truth answers, which is self-supervised.
Note that we build the MRC dataset using \textit{only} \texttt{SUP} samples, as the information in \texttt{REF} or \texttt{NEI} samples is indistinguishably untrustworthy and thus unable to be answered correctly.
During inference, the MRC model produces an answer $w_i' \in \mathcal{W}_E$ for a claim phrase $w_i \in \mathcal{W}_c$, which is used to replace $w_i$ for constructing a local premise.

A phrase in the claim may differ in surface form from the answers in the evidence, which is thus \textit{not} suitable for an \textit{extractive} MRC system.
Therefore, we adopt a \textit{generative} MRC model under the sequence-to-sequence (Seq2Seq) paradigm \cite{2020unifiedqa}.

\subsection{Veracity Prediction}

Given pre-computed local premises, we then use neural networks to parameterize $p_\theta(\bm{y}|\bm{z}, x)$ and the variational distribution $q_\phi(\bm{z}|\bm{y}, x)$ for veracity prediction.
They are optimized by the variational EM algorithm and decoded iteratively.

Given $c$, $E$ and local premises $\mathcal{P}$ for claim phrases respectively, we calculate the contextualized representations with pre-trained language models (PLMs).
We concatenate claim and each of the local premises with $\{x_\mathrm{local}^{(i)}=(c, c'_i)\}$ and encode them into hidden representations $\{\bm{h}^{(i)}_\mathrm{local} \in \mathbb{R}^d\}$.
Similarly, we encode the claim and concatenated evidence sentences as $x_\mathrm{global}=(c, E)$ into the global vector $\bm{h}_\mathrm{global} \in \mathbb{R}^d$, followed by a self-selecting module \cite{liu-etal-2020-multi} to find the important parts of a vector.

Not all phrases are the culprit phrase, so we design a \textit{culprit attention} based on a heuristic observation that: a valid local premise should be semantically close to the evidence sentences.
Thus, we design the similarity between $\bm{h}^{(i)}_\mathrm{local}$ and $\bm{h}_\mathrm{global}$ to determine the importance of the $i$-th claim phrase.
We calculate the context vector $\bm{h}_\mathrm{local}$ as follows:
\begin{equation}
    \label{eq:alpha}
    \bm{h}_\mathrm{local} = \tanh (\sum_{i=1}^{|\mathcal{W}_c|} \alpha_i \bm{h}^{(i)}_\mathrm{local});
    \alpha_i = \sigma(\mathbf{W_\alpha}[\bm{h}_\mathrm{global};\bm{h}^{(i)}_\mathrm{local}])
\end{equation}
where $\mathbf{W_\alpha} \in \mathbb{R}^{1 \times 2*d}$ is the parameter and $\sigma$ is the softmax function.

After calculating these representations, we design $p_\theta(\cdot)$ and $q_\phi(\cdot)$ both to be two-layer MLPs, where the last layer is shared as label embeddings:
\begin{itemize}
    \item $q_\phi(\bm{z}_i | \bm{y}, x)$ takes as input the concatenation of the label embeddings of $\bm{y}$ (ground truth $y^\star$ in training), $\bm{h}_\mathrm{local}^{(i)}$ and $\bm{h}_\mathrm{global}$, and outputs the probability of $\bm{z}_i$. 
    Note that $q_\phi(\bm{z} | \bm{y}, x) = \prod_i q_\phi(\bm{z}_i | \bm{y}, x)$.
    \item $p_\theta(\bm{y} | \bm{z}, x)$ takes as input the concatenation of $(\bm{z}_1, \bm{z}_2, \dots, \bm{z}_\mathrm{\max})$ ($\max$ length by padding),  $\bm{h}_\mathrm{global}$ and $\bm{h}_\mathrm{local}$, and outputs the distribution of $\bm{y}$.
\end{itemize}
During training, $q_\phi(\cdot)$ and $p_\theta(\cdot)$ are jointly optimized with Eq. \ref{eq:loss}.
We use the Gumbel reparameterization \cite{DBLP:conf/iclr/JangGP17} for discrete argmax operation from $\bm{z}$.
Specifically, we keep the argmax node and perform the usual forward computation (Gumbel Max),
but backpropagate a surrogate gradient (gradient of Gumbel Softmax).

\subsubsection{Decoding}

During inference, we \emph{randomly} initialize $\bm{z}$, and then iteratively decode $\bm{y}$ and $\bm{z}$ with $p_\theta (\bm{y}|\bm{z}, x)$ and $q_\phi (\bm{z}|\bm{y}, x)$ until convergence.
In the end, we have both the final prediction $y$ and the latent variables $\bm{z}$ serving as the phrasal veracity predictions for all claim phrases.

%% file: 030experiment.tex
\subsection{Dataset and Evaluation Metrics}
\label{sec:metric}

\subsubsection{Dataset}

We evaluate our verification methods on a large-scale fact verification benchmark, i.e., FEVER 1.0 shared task \cite{thorne-etal-2018-fever}, which is split into \textit{training}, \textit{development} and \textit{blind test} set.
FEVER utilizes Wikipedia (dated June 2017) as the trustworthy knowledge source from which the evidence sentences are extracted.
The statistical report of FEVER dataset is presented in Table \ref{tab:dataset}, with the split sizes of SUPPORTED (\texttt{SUP}), REFUTED (\texttt{REF}) and NOT ENOUGH INFO (\texttt{NEI}) classes.
In this dataset, there are 3.3 phrases per claim/question on average.

\begin{table}[t]
    \centering
    \begin{tabular}{cccc}
        \toprule
         & \textbf{Training} & \textbf{Development} & \textbf{Test} \\
        \midrule
        \texttt{SUP} & 80,035 & 6,666 & 6,666 \\
        \texttt{REF} & 29,775 & 6,666 & 6,666 \\
        \texttt{NEI} & 35,659 & 6,666 & 6,666 \\
        \bottomrule
    \end{tabular}
    \caption{Statistics of FEVER 1.0 dataset.}
    \label{tab:dataset}
\end{table}

\subsubsection{Evaluation Metrics}
Following previous studies, we evaluate the systems using: 
\begin{itemize}
    \item \textit{\textbf{L}abel \textbf{A}ccuracy} (LA): The accuracy of predicted label for claim regardless of retrieved evidence; 
    \item \textit{\textbf{FEV}ER score} (FEV): The accuracy of both predicted label and retrieved evidence, which encourages the correct prediction based on correct retrieved evidence.
\end{itemize}
In other words, FEVER score rewards a system that makes predictions based on correct evidence.
Note that no evidence is needed if a claim is labeled \texttt{NEI}.

In addition, we propose several metrics to evaluate the quality of explanations, i.e., phrasal veracity predictions $z$:
\begin{itemize}
    \item \textit{Logically aggregated \textbf{L}abel \textbf{A}ccuracy} (LA$_z$): We calculate the accuracy of logically aggregated $y_z$ by Eq. \ref{eq:softlogic}, which evaluates the \emph{overall} quality of explanations $z$;
    \item \textit{\textbf{Culp}rit finding \textbf{A}bility} (\textsc{CulpA}): 
    LA$_z$ cannot evaluate \textit{individual} phrase veracity $z_i$ or decide whether a model finds the correct culprit phrase.
    Thus, we randomly select 100 refuted claims from development set, and manually label the culprit phrases (allowing multiple culprits).\footnote{Note that the set of annotated culprits is a subset of the extracted claim phrases for the convenience of calculation. We find that there are on average 1.26 culprit phrases per claim for the sampled ones, indicating that the refuted claims in the FEVER dataset generally have a single culprit.}
    \textsc{CulpA} calculates the \textbf{P}recision, \textbf{R}ecall and \textbf{F1} of the culprit finding based on \textit{discrete} veracity from $z$.
    \item \textit{\textbf{Agree}ment} (\textsc{Agree}): The agreement between predictions of aggregated veracity $y_z$ and the final veracity $y$, which evaluates the \emph{faithfulness} of explanations;
\end{itemize}
We use two ways of aggregation logic for calculating LA$_z$ and \textsc{Agree}, i.e., discrete \textit{hard} logic (as in $\mathsection$\ref{sec:logic}) and probabilistic \textit{soft} logic (as in $\mathsection$\ref{sec:prior}).

\subsection{Baseline Methods}

We evaluate \method against several public state-of-the-art baselines:
\begin{itemize}
    \item \textbf{UNC NLP} \cite{Nie2019CombiningFE} is the champion system in the FEVER competition, which uses ESIM \cite{chen-etal-2017-enhanced} to encode pairs of claim and evidence sentence, enhanced with internal semantic relatedness scores and WordNet features.
    \item \textbf{GEAR} \cite{zhou-etal-2019-gear}, which is a pioneer model to utilize BERT \cite{devlin-etal-2019-bert} to model the interaction between claim and evidence sentence pairs, followed by a graph network for the final prediction. 
    \item \textbf{DREAM} \cite{zhong2020reasoning}, which is built on top of an XLNet \cite{yang2019xlnet} and breaks the sentences into semantic graphs using semantic role labeler, followed by a graph convolutional network \cite{velickovic2018graph} and graph attention for propagation and aggregation.
    \item \textbf{KGAT} \cite{liu2020fine}, which collapses sentences into nodes, encodes them with RoBERTa \cite{liu2019roberta}, and adopts a Kernel Graph Attention Network for aggregation. 
    Further research equips KGAT with CorefRoBERTa \cite{ye-etal-2020-coreferential}, a PLM designed to capture the relations between co-referring noun phrases.
    \item \textbf{LisT5} \cite{jiang-etal-2021-exploring-listwise} is currently the champion in FEVER 1.0 shared tasks. 
    LisT5 employs a list-wise approach with data augmentation on top of a T5-3B \cite{raffel2020exploring} with 3 billion parameters, which is almost 10 times larger than the large versions of BERT, RoBERTa and XLNet.
\end{itemize}
We note that the comparison between baselines is not always fair due to too many different settings such as evidence retrieval and backbone pre-trained language models.

\subsection{Implementation Details}
\label{sec:further_impl}

We describe the implementation details in the experiments for the following.
\method consists of a pipeline of modules, among which the MRC model and the verification model are trained by exploiting the FEVER dataset.
All of the backbone PLMs inherit HuggingFace's implementation \cite{wolf-etal-2020-transformers} as well as most of the parameters.

\paragraph{Training Details of MRC}
We train the model in a self-supervised way, i.e., using the \texttt{SUP} samples in training and development set.
The constructed dataset consists of 80,035 training samples and 6,666 development samples, corresponding to the statistics of \texttt{SUP} samples in Table \ref{tab:dataset}.

We fine-tune a $\text{BART}_{base}$ model \cite{lewis-etal-2020-bart} for the MRC model.
Following the standard Seq2Seq training setup, we optimize the model with cross entropy loss.
We apply AdamW as the optimizer during training.
We train the model for 4 epochs with initial learning rate of 5e-5, and use the checkpoint with the best ROUGE-2 score on the development set.

\paragraph{Training Details of Veracity Prediction}
During data pre-processing, we set the maximum lengths of $x_\mathrm{global}$ and $x^{(i)}_\mathrm{local}$ as 256 and 128 tokens respectively, and set the maximum number of phrases per claim as 8. 
For each claim phrase $w_i$, we keep the top 3 answers in the beam search as candidates from the MRC model, replace $w_i$ with them respectively, and concatenate the sentences as the local premise for the claim phrase $w_i$.
During training, we set the initial learning rate of \method with BERT and RoBERTa as 2e-5 and 1e-5, and batch size as 16 and 8 respectively. 
The models are trained on 4 NVIDIA Tesla V100 GPUs for $\sim$5 hours for best performance on development set. 
We keep checkpoints with the highest label accuracy on the development set for testing. 
During inference, decoding quickly converges after 2 or 3 iterations.

\paragraph{Evidence Retrieval} 
Since the primary focus of this work is fact verification, we directly adopt the evidence retrieval methods from KGAT \cite{liu2020fine} for comparison in the verification sub-task.
We leave the reported performance of several evidence retrieval techniques and the results of \method with oracle evidence retrieval in Appendix. 

%% file: 040results.tex
In this section, we evaluate the performance of \method compared with baselines and analyze the interpretability of \method w.r.t. phrase veracity and local premise quality.\footnote{We set $\lambda=0.5$ by default in our experiments.}

\subsection{Overall Performance}

\begin{table}[tb]
    \centering
    \begin{tabular}{lcccc}
        \toprule
        \multirow{2}{*}{\textbf{Model}} & \multicolumn{2}{c}{\textbf{Dev}} & \multicolumn{2}{c}{\textbf{Test}} \\
        \cmidrule(lr){2-3}
        \cmidrule(lr){4-5}
        &\textbf{LA} & \textbf{FEV} & \textbf{LA} & \textbf{FEV} \\ 
        \midrule 
        UNC NLP & 69.72 & 66.49 & 68.21 & 64.21 \\
        GEAR (BERT$_\mathrm{base}$) & 74.84 & 70.69 & 71.60 & 67.10 \\
        \midrule
        DREAM (XLNet$_\mathrm{large}$) & 79.16 & - & \underline{76.85} & 70.60 \\
        KGAT (BERT$_\mathrm{large}$) & 77.91 & 75.86 & 73.61 & 70.24 \\
        \ \ \ \ $\llcorner$ (RoBERTa$_\mathrm{large}$) & 78.29 & 76.11 & 74.07 & 70.38 \\
        \ \ \ \ $\llcorner$ (CorefRoBERTa$_\mathrm{l}$) & - & - & 75.96 & 72.30 \\
        \method (BERT$_\mathrm{large}$) & 78.44 & 76.21 & 74.43 & 70.71 \\
        \ \ \ \ $\llcorner$ (RoBERTa$_\mathrm{large}$) & \underline{81.14} & \textbf{78.83} & 76.42 & \underline{72.93} \\
        \midrule
        LisT5 (T5$_\mathrm{3B}$) & \textbf{81.26} & \underline{77.75} & \textbf{79.35} & \textbf{75.87} \\
        \bottomrule
    \end{tabular}
    \caption{Overall performance of verification results on the dev and blind test set of FEVER task, where FEV (FEVER score) is the main evaluation metric. The best is \textbf{bolded}, and the second best is \underline{underlined}.}
    \label{tab:main}
\end{table}

Table \ref{tab:main} reports the overall performance of \method compared with baselines on the development and test set of FEVER. 
In general, \method outperforms or is comparable to published baseline methods of similar sizes.
LisT5 shows its superiority over other methods, which may be mainly attributed to its much larger and more powerful PLM (T5-3B).
Still, \method outperforms LisT5 in FEV score in the development set.
For DREAM, we notice it achieves better score in LA score in the test set than \method.
Due to the difference in evidence retrieval strategies and backbone PLMs, \method is not fully comparable with DREAM.
However, a higher FEV score of \method (for both BERT and RoBERTa) indicates it makes decisions more faithful to evidence than DREAM.
In contrast, we make fairer comparisons with KGAT (same PLMs and evidence retrieval techniques), and find that \method with BERT$_\mathrm{large}$ and RoBERTa$_\mathrm{large}$ beats KGAT with RoBERTa$_\mathrm{large}$ and CorefRoBERTa$_\mathrm{large}$, respectively.

We then perform a detailed analysis of the proposed components in \method (RoBERTa$_\mathrm{large}$) on the development set to assess their influences on the performance and explanation quality.


\begin{table}[t]
    \centering
    \begin{tabular}{cccccc}
    \toprule
        \multirow{2}{*}{$\lambda$ in $\mathcal{L}_\mathrm{final}$} & \multirow{2}{*}{\textbf{LA}} & \multicolumn{2}{c}{\textbf{LA$_z$}} & \multicolumn{2}{c}{\textbf{\textsc{Agree}}}\\
        \cmidrule(lr){3-4}
        \cmidrule(lr){5-6}
        & & Hard & Soft & Hard & Soft \\
        \midrule
        $\lambda = 0.0$ & 81.10 & 51.99 & 51.92 & 54.02 & 53.90  \\
        $\lambda = 0.3$ & 80.98 & 75.24 & 78.75 & 90.06 & 93.14  \\
        $\lambda = 0.5$ & \textbf{81.14} & 76.54 & 79.66 & 92.94 & 96.11  \\
        $\lambda = 0.7$ & 80.92 & \textbf{77.77} & \textbf{80.28} & \textbf{93.81} & 96.79  \\
        $\lambda = 0.9$ & 80.28 & 75.55 & 80.02 & 91.56 & \textbf{98.43}  \\
    \bottomrule
    \end{tabular}
    \caption{Evaluation of phrase veracity quality with the adjustment of the balancing weight $\lambda$ in the loss function (Eq. \ref{eq:loss}) in \method. 
    The accuracy and faithfulness of phrase veracity boost as $\lambda$ increases towards $\mathcal{L}_\mathrm{logic}$.}
    \label{tab:lambda}
\end{table}

\subsection{Evaluation of Phrase Veracity}

As one of the most important features in \method, phrasal veracity predictions explain the verification results.
Therefore, such explanations must be accurate as well as faithful to the final prediction.
Since the hyper-parameter $\lambda$ controls the influence of logical constraints, we perform an ablation study of $\lambda$, where $\lambda=0$ indicates no logical constraints on the latent variables.

As seen in Table \ref{tab:lambda}, we report the results of LA, LA$_z$ and \textsc{Agree} which comprehensively evaluate the general quality of phrasal veracity predictions.
We have three major observations from the table: 
\begin{inparaenum}[\it 1)]
    \item Aggregation with soft logic is better than hard logic in terms of accuracy and faithfulness.
    This indicates that predicted probability distributions of phrase veracity are important and gives more information than discrete labels.
    \item In general, the explanations are faithful, with over 96\% of aggregated phrase veracity consistent with the claim veracity. The explanations are also accurate according to LA$_z$ and LA scores. 
    \item With the increase of $\lambda$ and stronger regularization of $\mathcal{L}_\mathrm{logic}$, the general accuracy and faithfulness of phrase veracity increase.
    Without $\mathcal{L}_\mathrm{logic}$, \method cannot give any meaningful explanations.
\end{inparaenum}

In summary, the results demonstrate the effectiveness of phrase veracity and the importance of the aggregation logic.

\paragraph{Ablation on Prior Distribution}
\label{sec:priorablation}

\begin{table}[t]
    \centering
    \small
    \begin{tabular}{lcccc}
        \toprule
        \textbf{Choice of $p(z)$} & \textbf{LA} & \textbf{LA$_z$} & \textbf{\textsc{Agree}} & \textbf{\textsc{CulpA} (P/R/F1)} \\
        \midrule
        NLI prior & \textbf{81.14} & 79.66 & 96.11 & \textbf{75.8}/75.9/\textbf{74.3} \\
        Pseudo prior & 80.93 & 80.44 & \textbf{97.25} & 70.5/77.1/71.4 \\
        Uniform prior & 80.85 & \textbf{80.74} & 97.08 & 34.1/\textbf{78.8}/46.1 \\
        \bottomrule
    \end{tabular}
    \caption{Results of different choices of prior distribution $p(z)$ during training, where $y_z$ in LA$_z$ is calculated using \textit{soft} logic. 
    }
    \label{tab:kd}
\end{table}

As presented in $\mathsection$\ref{sec:prior}, we use the results of a fixed, off-the-shelf NLI model \cite{he2021deberta} as the prior distribution $p(\bm{z})$.
We first evaluate the quality of NLI predictions in this task by directly making them as phrasal veracity predictions.
We make local premises as \textit{premise} and the claim as \textit{hypothesis}.
The predictions are aggregated into $y_z$ using the same soft logic, and we get the LA$_z$ score at only 53.41\%.
However, with \method training, LA$_z$ can reach the score at 79.66\% or more.

We further perform an ablation study to investigate the influence of the choice of prior distribution.
We propose two alternatives:
\begin{enumerate}
    \item \textit{logical pseudo distribution}. We create pseudo $p(\bm{z})$ and sample 1 or 2 phrases as the culprit phrase(s) based on culprit attention weight $\alpha$ in Eq. \ref{eq:alpha}, and label them as \texttt{REF} and the rest as non-\texttt{REF}. Such $p(\bm{z})$ is in accordance with the logic but distinguishable of the culprit phrase(s);
    \item \textit{uniform distribution}, which is commonly used as $p(\bm{z})$.
\end{enumerate}
$z$ is \emph{randomly} initialized during decoding in all scenarios.

As reported in Table \ref{tab:kd}, after switching prior distributions, the model still performs well and learns logically consistent phrase veracity w.r.t. LA, LA$_z$ and \textsc{Agree}.
For logical pseudo prior, LA$_z$ and \textsc{Agree} are better than NLI prior since there is gap between off-the-shelf NLI models and this task.
But their scores on \textsc{CulpA} are close, proving similar culprit finding ability for both prior distributions.
However, with uniform distribution, \method makes the \textit{same} predictions for all claim phrases, which results in high \textsc{CulpA} recall (78.8\%) but poor F1 scores due to its indistinguishability.

\subsection{Evaluation of MRC Quality}

\begin{table}[t]
    \centering
    \begin{tabular}{lccc}
    \toprule
        \multirow{2}{*}{\textbf{MRC Model}} & \multicolumn{3}{c}{\textbf{MRC Acc}}\\
        \cmidrule{2-4}
        & \texttt{SUP} & \texttt{REF} & \texttt{NEI} \\
    \midrule
    UnifiedQA (hit@1) & 43.90 & 39.47 & 30.00 \\
    UnifiedQA (hit@4) & 56.10 & 52.63 & 47.50 \\
    \method (hit@1) & \textbf{95.12} & 78.95 & 83.75 \\
    \method (hit@4) & \textbf{95.12} & \textbf{89.47} & \textbf{87.50} \\
    \bottomrule
    \end{tabular}
    \caption{Manual evaluation of the performance of MRC models. Hit@$k$ denotes that we keep the top-$k$ answers in the beam search as the candidates. The answer is accurate if any one of the $k$ answers is correct. 
    }
    \label{tab:qa}
\end{table}

The system should acquire enough distinguishable information to know the veracity of claim phrases.
One of the key designs in \method is using an MRC model to retrieve evidential phrases for verifying claim phrases, i.e., constructing local premises.
In this section, we evaluate the quality of the MRC model and its influence on culprit finding.

Since there are no ground truth answers for \texttt{REF} and \texttt{NEI} claims, we \textit{manually} evaluate the MRC model in \method, which is a BART$_\mathrm{base}$ fine-tuned in a self-supervised way.
The data samples are labeled as \textit{correct} if they are the right answer(s) for verifying the claim phrase, otherwise \textit{erroneous}.\footnote{We extract the correct answers from evidence manually for evaluation. 
For \texttt{NEI} samples, there could be some claim phrases that do not have correct counterparts in the evidence. 
So we decide the MRC results for those phrases to be correct if the results are the same as claim phrases.}
We randomly selected 20 data samples per class for manual evaluation, with a total of 60 samples and 238 QA pairs (also 238 claim phrases).
As a zero-shot baseline, we adopt UnifiedQA \cite{2020unifiedqa}, which fine-tunes a T5$_\mathrm{base}$ \cite{raffel2020exploring} on existing QA tasks.

Results in Table \ref{tab:qa} reveal the effectiveness of self-supervised training for adaptation and room for future refinement.
Note that, different from traditional MRC tasks, the question can contain false information for non-\texttt{SUP} cases. 
Thus the accuracy drops as the question deteriorates.
The results shed light on the \textit{automatic correction} while performing verification \cite{thorne-vlachos-2021-evidence} since the answers can serve as a correction proposal.

\paragraph{Influence of MRC Performance}

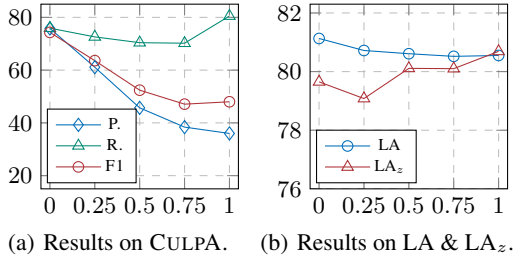
\begin{figure}[t]
    \centering
	\subfigure[Results on \textsc{CulpA}.] { \label{fig:culpa} 
		\input{figs/maskculpa}}
	\subfigure[Results on LA \& LA$_z$.] { 
	\label{fig:la} 
		\input{figs/maskla}}
    \caption{Performance on culprit finding (\textsc{CulpA}) and verification (LA and LA$_z$) vs. the mask rate $\rho$ of local premises, simulating the influence by deficiency of the MRC model.}
    \label{fig:mask}
\end{figure}

We further analyze the influence brought by the quality of the MRC model.
To do so, we randomly mask the local premises at the rate of $\rho$ (e.g., \textit{Donald Trump won \texttt{[MASK]}}.), simulating the failure of the MRC model in an extreme situation.
As seen in Figure \ref{fig:mask}, in general, the quality of local premises are critical for identifying the culprit phrases.
In Figure \ref{fig:culpa}, F1 score of \textsc{CulpA} quickly deteriorates as the quality of local premises gets worse.
When mask rate reaches 100\%, precision drops to 36.0\% but recall hits 80.5\%.
This is because \method no longer identifies the culprit phrase and predicts all phrases to be the same, which is similar to the scenario of uniform prior distribution as discussed in $\mathsection$\ref{sec:priorablation}.
From Figure \ref{fig:la}, we find claim verification ability (LA) of \method does not drop much, which is partly because the answers are already displayed in the evidence text.
Also, the gap between LA$_z$ and LA gradually narrows as mask rate ascends, because phrase verification degenerates into claim verification and makes the same predictions when local premises do not provide targeted information for claim phrases.

\subsection{Case Study}

\begin{figure}[t]
    \centering
    \includegraphics[width=\linewidth]{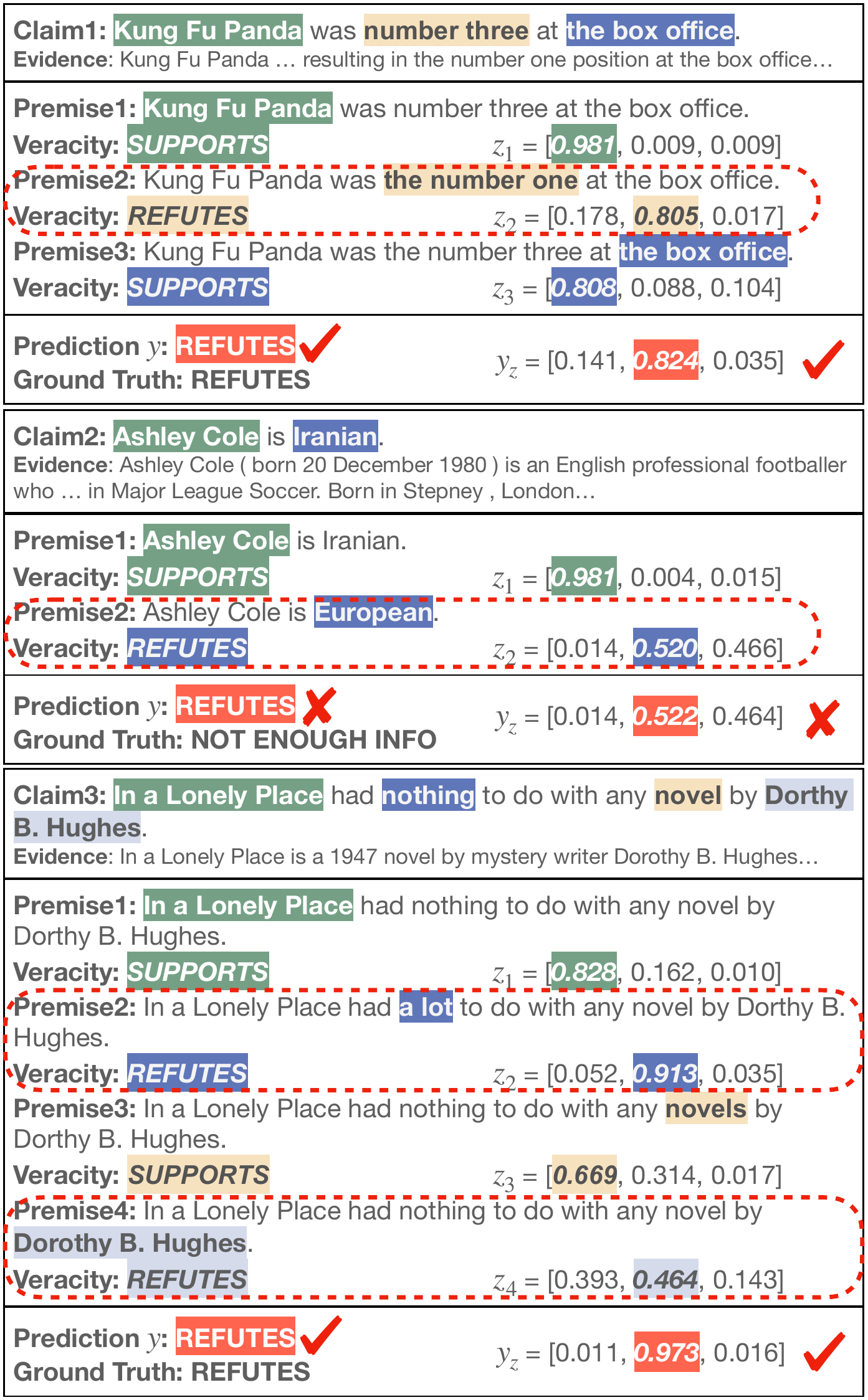}
    \caption{A case study of the interpretability of \method, where the probabilities in phrasal veracity prediction $z_i$ are \texttt{SUP}, \texttt{REF} and \texttt{NEI} respectively.}
    \label{fig:case1}
\end{figure}

We present three examples in Figure \ref{fig:case1} to show the interpretability of \method.
In the first, \method performs well in both claim and phrasal veracity predictions.
It successfully finds the culprit phrase ``\textit{number three}'', and a correction suggestion by MRC, i.e., ``\textit{number one}'' in Premise 2.

In the second example, \method makes mistakes by predicting the veracity of the second phrase to be \texttt{REF}.
The root causes for this mistake are complicated, including lack of commonsense knowledge and failure of the MRC and evidence retrieval modules.
The MRC retrieves ``\textit{European}'' (hit@1) for filling the masked ``\textit{Iranian}'', whereas there is no definite answer to be drawn from the evidence.
Strictly speaking, we can only know from the evidence text that \textit{Ashley Cole} was born in England, but do not know whether he has dual citizenship or joined another country for certain.
Therefore, we have \textit{not enough information} (\texttt{NEI}) to draw the verdict, but \method predict it to be \texttt{REF}.
However, the probability of \texttt{NEI} and \texttt{REF} for phrasal veracity prediction $z_2$ (0.466 vs. 0.520) and for claim veracity $y_z$ (0.464 vs. 0.522) are rather close, which indicates that \method struggles to make that decision.
These findings stress the usefulness and interpretability of the predicted phrase veracity $z$.

We investigate a multiple culprits scenario in the third example.
The last three phrases in claim 3 could be seen as the culprits, and \method predicts ``\textit{nothing}'' and ``\textit{Dorthy B. Hughes}'' as \texttt{REF}.
This corroborates that \method is by design capable of detecting multiple culprits in a claim.

%% file: figs/maskculpa.tex


\pgfplotsset{width=0.5\linewidth,height=0.48\linewidth,compat=1.7}
\footnotesize
\begin{tikzpicture}
\begin{axis}[
    xmin=-0.05, xmax=1.05,
    ymin=15, ymax=85,
    xtick={0.0, 0.25, 0.5, 0.75, 1.0},
    ytick={20, 40, 60, 80},
    legend pos=south west,
    ymajorgrids=true,
    xmajorgrids=true,
    grid style=dashed,
    x label style={at={(axis description cs:0.5,-0.125)},anchor=north},
    y label style={at={(axis description cs:-0.125,0.5)},anchor=south},
    legend style={nodes={scale=0.7, transform shape}}
]
\addplot[
    color=NavyBlue,
    mark=diamond,
    mark size=2.5pt,
    ]
    coordinates {
    (0.0, 75.8)
    (0.25, 61.1)
    (0.5, 45.7)
    (0.75, 38.4)
    (1.0, 36.0)
    };
    \addlegendentry{\textsc{P.}}

\addplot[
    color=PineGreen,
    mark=triangle,
    mark size=2.5pt,
    ]
    coordinates {
    (0.0, 75.9)
    (0.25, 72.6)
    (0.5, 70.4)
    (0.75, 70.2)
    (1.0, 80.5)
    };
    \addlegendentry{\textsc{R.}}

\addplot[
    color=Maroon,
    mark=o,
    mark size=2pt,
    ]
    coordinates {
    (0.0, 74.3)
    (0.25, 63.6)
    (0.5, 52.4)
    (0.75, 47.1)
    (1.0, 48.0)
    };
    \addlegendentry{\textsc{F1}}

\end{axis}
\end{tikzpicture}


%% file: figs/maskla.tex
\pgfplotsset{width=0.5\linewidth,height=0.48\linewidth,compat=1.5}
\footnotesize
\begin{tikzpicture}
\begin{axis}[
    xmin=-0.05, xmax=1.05,
    ymin=76, ymax=82.3,
    xtick={0.0, 0.25, 0.5, 0.75, 1.0},
    legend pos=south west,
    ymajorgrids=true,
    xmajorgrids=true,
    grid style=dashed,
    x label style={at={(axis description cs:0.5,-0.125)},anchor=north},
    y label style={at={(axis description cs:-0.125,0.5)},anchor=south},
    legend style={nodes={scale=0.7, transform shape}}
]
\addplot[
    color=NavyBlue,
    mark=o,
    mark size=2pt,
    ]
    coordinates {
    (0.0, 81.13)
    (0.25, 80.72)
    (0.5, 80.61)
    (0.75, 80.52)
    (1.0, 80.55)
    };
    \addlegendentry{LA}
\addplot[
    color=Maroon,
    mark=triangle,
    mark size=2.5pt,
    ]
    coordinates {
    (0.0, 79.65)
    (0.25, 79.08)
    (0.5, 80.11)
    (0.75, 80.10)
    (1.0, 80.69)
    };
    \addlegendentry{LA$_z$}
\end{axis}
\end{tikzpicture}

%% file: 060conclusion.tex
In this paper, we propose \method, an approach for interpretable fact verification by distilling the logical knowledge into the latent model.
In the experiments, we find \method not only enjoys competitive performance with baselines but produces faithful and accurate phrase veracity predictions as explanations.
Besides, the local premises constructed by the self-supervised MRC module are of high quality and deeply influence the finding of culprits, making \method's ability of automatic factual correction worthy of investigation in the future.

We add that, a general notion of culpability discovery in fact verification may depend on claim decomposition.
A claim should be decomposed into fine-grained units where the culprit hides while making the units self-explanatory to humans.
Besides phrases introduced in this paper, there could be other forms of decomposition units, e.g., dependency arc.
We suggest future research focus on the limitations of \method, including decomposition, evidence retrieval, and out-of-domain issues.
Accordingly, better solutions for these issues can improve \method's generality.

%% file: 070supplementary.tex
\section{Evidence Retrieval Results}
\label{sec:er}

For the sake of completeness, we describe here the commonly adopted method for evidence retrieval. 
Given a claim sentence, the evidence retrieval system first identifies entity phrases in the sentence and then searches for Wikipedia pages with the given entity names (e.g., \textit{Donald Trump}).
It further selects related sentences from the retrieved Wikipedia pages based on a neural sentence ranking model.

Since we focus on the second sub-task, we here describe the evidence retrieval techniques adopted in baselines.
To illustrate the effectiveness of these methods, we present the reported results of these methods in Table \ref{tab:ev} based on the top-5 retrieved evidence sentences per claim. 
According to Table \ref{tab:ev}, ER-KGAT, and ER-DREAM are consistently better than ER-ESIM, and they are comparable on the blind test set, yet ER-KGAT outperforms ER-DREAM on the development set.
By default, we use evidence retrieved using ER-KGAT in \method for the following experiments.

\begin{table}[htp]
    \centering
    \begin{tabular}{lcccc}
        \toprule
        & \textbf{ER Method} & \textbf{Prec@5} & \textbf{Rec@5} & \textbf{F1@5} \\
        \midrule
        \multirow{3}{*}{\textbf{Dev}} & ER-ESIM & 24.08 & 86.72 & 37.69 \\
        & ER-DREAM & 26.67 & 87.64 & 40.90  \\
        & ER-KGAT & \textbf{27.29} & \textbf{94.37} & \textbf{42.34} \\
        \midrule 
        \multirow{3}{*}{\textbf{Test}} & ER-ESIM & 23.51 & 84.66 & 36.80 \\
        & ER-DREAM & \textbf{25.63} & 85.57 & \textbf{39.45} \\
        & ER-KGAT & 25.21 & \textbf{87.47} & 39.14 \\
        \bottomrule
    \end{tabular}
    \caption{Reported performance of evidence retrieval (ER) strategies on \textbf{Prec}ision@5, \textbf{Rec}all@5 and \textbf{F1}@5. }
    \label{tab:ev}
\end{table}

\subsubsection{Performance with Oracle Evidence Retrieval}

Recall that \method focuses on the fact verification sub-task and uses evidence sentences retrieved from another system.
Even with the relatively good MRC module as reported in Table \ref{tab:qa}, the \method pipeline still suffers from the initial evidence retrieval error.
What if \method uses perfect evidence without information losses from evidence retrieval system?
To answer this question, we fill claims in the development set with ground truth evidence sentences and excluding the 1/3 \texttt{NEI} cases since they do not have oracle evidence.
Results in Table \ref{tab:oracle} show somehow the upper bound of \method, which also suggests a viable direction for future improvements.

\begin{table}[ht]
    \centering
    \begin{tabular}{lcc}
        \toprule
        \method & \textbf{LA} & \textbf{FEV} \\
        \midrule
        w/ ER-KGAT & 86.12 & 82.66 \\
        w/ ER-Oracle & \textbf{88.92} & \textbf{88.62} \\
        \bottomrule
    \end{tabular}
    \caption{Verification results of \method using retrieved evidence (ER-KGAT) and oracle evidence (ER-Oracle). We exclude \texttt{NEI} cases since they do not have corresponding evidence.}
    \label{tab:oracle}
\end{table}